\documentclass{article}

     \usepackage[nonatbib]{neurips_2020}

\usepackage[utf8]{inputenc} 
\usepackage[T1]{fontenc}    
\usepackage{hyperref}       
\usepackage{url}            
\usepackage{booktabs}       
\usepackage{amsfonts}       
\usepackage{nicefrac}       
\usepackage{microtype}      
\usepackage{mathrsfs}
\usepackage{graphicx}
\usepackage{subfigure}
\usepackage{amsmath}
\usepackage{booktabs}

\usepackage[noend]{algpseudocode}

\usepackage{algorithmicx,algorithm}

\title{Ordinal Pattern Kernel for Brain Connectivity Network Classification}

\author{
  Kai Ma\\
  \texttt{kaim@nuaa.edu.cn}
  \And
  Biao Jie \\
  \texttt{jbiao@nuaa.edu.cn}
  \And
  Daoqiang Zhang\thanks{Corresponding author} \\
  \texttt{dqzhang@nuaa.edu.cn}
  \\
  Department of Computer Science and Technology\\
  MIIT Key Laboratory of Pattern Analysis and Machine Intelligence\\
  Nanjing University of Aeronautics and Astronautics\\
  Jiangsu Province, China, 210016 \\
}

\begin{document}

\maketitle

\begin{abstract}
    Brain connectivity networks, which characterize the functional or structural interaction of brain regions, has been widely used for brain disease classification. Kernel-based method, such as graph kernel (i.e., kernel defined on graphs), has been proposed for measuring the similarity of brain networks, and yields the promising classification performance. However, most of graph kernels are built on unweighted graph (i.e., network) with edge present or not, and neglecting the valuable weight information of edges in brain connectivity network, with edge weights conveying the strengths of temporal correlation or fiber connection between brain regions. Accordingly, in this paper, we present an ordinal pattern kernel for brain connectivity network classification. Different with existing graph kernels that measures the topological similarity of unweighted graphs, the proposed ordinal pattern kernels calculate the similarity of weighted networks by comparing ordinal patterns from weighted networks.
    To evaluate the effectiveness of the proposed ordinal kernel, we further develop a depth-first-based ordinal pattern kernel, and perform extensive experiments in a real dataset of brain disease from ADNI database. The results demonstrate that our proposed ordinal pattern kernel can achieve better classification performance compared with state-of-the-art graph kernels.
\end{abstract}

\section{Introduction}

Brain connectivity network characterizes the abstract connection structure of human brain, where brain regions correspond to nodes and functional or anatomical associations between nodes are considered as edges. Brain network is widely applied to classification of brain diseases, including Alzheimer’s disease (AD) [1], attention deficit hyperactivity disorder (ADHD) [2], major depressive disorder (MDD) [3] and schizophrenia [4]. In these studies, various network measures, e.g., degree, clustering coefficient [5, 6], are first extracted from connectivity networks as features for classification.

The graph kernels, which measure the topological similarity of brain network, have shown promising performance on all kinds of classification problems[7,8]. There are a variety of graph kernels that are different from each other in topological structures, including paths, walks, trees and subgraphs. Shortest-path kernel [9] is a graph kernel based on paths. Random walk graph kernels [10, 11] and return probability graph kernel [12] are the graph kernels based on walks. Cyclic pattern kernels [13, 14], tree pattern kernels [15, 16], Weisfeiler-Lehman graph kernel [17] and its variant [18] are the graph kernels based on trees. Subgraph matching kernels [19] are based on subgraphs. Pyramid Match kernel [20, 21] is based on pyramid structure. These graph kernels are widely used to classify the network structured data, such as molecules. But these graph kernels are defined on unweighted graph with edge present or not, thus neglect the valuable weight information of edges in brain connectivity network, with edge weights conveying the strengths of temporal correlation or fiber connection between brain regions.

To address this problem, we develop an ordinal pattern kernel for measuring the brain network similarities. In this work, we firstly introduce our proposed ordinal pattern kernels and provide the theoretical foundations for them. Then, we find that computing ordinal pattern kernels is NP-hard. In order to avoid the NP-hard problem in ordinal pattern kernels, we propose depth-first-based ordinal pattern kernel. At last, we perform the classification experiments and ordinal sub-structure mining experiments in the network data of brain diseases. Specifically, our work has following advantages:
%
\begin{itemize}\setlength{\itemsep}{0pt }
\item Our ordinal pattern kernel could make full use of weight information of edge in brain network and outperforms the existing state-of-the-art graph kernels in the classification accuracy.
\item Our ordinal pattern kernel has strong robustness. When brain networks have missing data, our method could still acquire the best classification accuracy.
\item Our proposed depth-first-based ordinal pattern could capture the discriminative sub-structures for seeking the biomarkers in brain disease.
\end{itemize}


\section{Background}
\subsection{Graph kernels}
Graph kernels are a class of kernel functions measuring the similarities between graphs. There is a map $\phi$, which could implicitly embed the original graph data set $\mathcal{G}$ into a Hilbert space $\mathcal{H}$, $\phi$: $\mathcal{G}$ $\rightarrow$ $\mathcal{H}$. In $\mathcal{G}$, graph kernel $\mathcal{K}$: $\mathcal{G}$ $\times$ $\mathcal{G}$ $\rightarrow$ $\mathbb{R}$ is a function associated with $\mathcal{H}$, given two graphs \emph{G$_1$} and \emph{G$_2$}, \emph{G$_1$}, \emph{G$_2$} $\in$ $\mathcal{G}$, graph kernel $\mathcal{K}$ is interpreted as a dot product in the high dimensional space $\mathcal{H}$, $\mathcal{K}\left(G_1,G_2\right)=\langle\phi\left(G_1\right),\phi\left(G_2\right)\rangle_\mathcal{H}$. If $\mathcal{H}$ is a reproducing kernel Hilbert space (RKHS), then $\mathcal{K}$ is a positive definite kernel.

\subsection{Isomorphism}
\emph{A} and \emph{B} are two nonempty sets, $\varphi$ is a map from \emph{A} to \emph{B}. $\circ$ and $\bar{\circ}$ are respectively the algebraic operation on \emph{A} and \emph{B}. If $\forall\emph{a}, \emph{b}\in\emph{A}$, then $\varphi\left(\emph{a}\circ\emph{b}\right)=\varphi\left(\emph{a}\right) \bar{\circ} ~\varphi\left(\emph{b}\right)$, $\varphi$ is called the homomorphic mapping from \emph{A} to \emph{B}. If $\varphi$ is a homomorphic and onto mapping from \emph{A} to \emph{B}, we call \emph{A} and \emph{B} are homomorphic. If $\varphi$ is a homomorphic and bijective mapping from \emph{A} to \emph{B}, then $\varphi$ is called the isomorphic mapping from \emph{A} to \emph{B}. We call \emph{A} and \emph{B} are isomorphic, $\emph{A} \cong \emph{B}$.

\section{Ordinal pattern}
Ordinal pattern  is regarded as a new descriptor for brain connectivity networks [22], which provides ordinal edge sequences for each node. Here, we extend ordinal pattern into graph and redefine it with graph theory.
\subsection{Ordinal pattern}
\begin{figure}
\centering
\includegraphics[width=11cm,height=4.2cm]{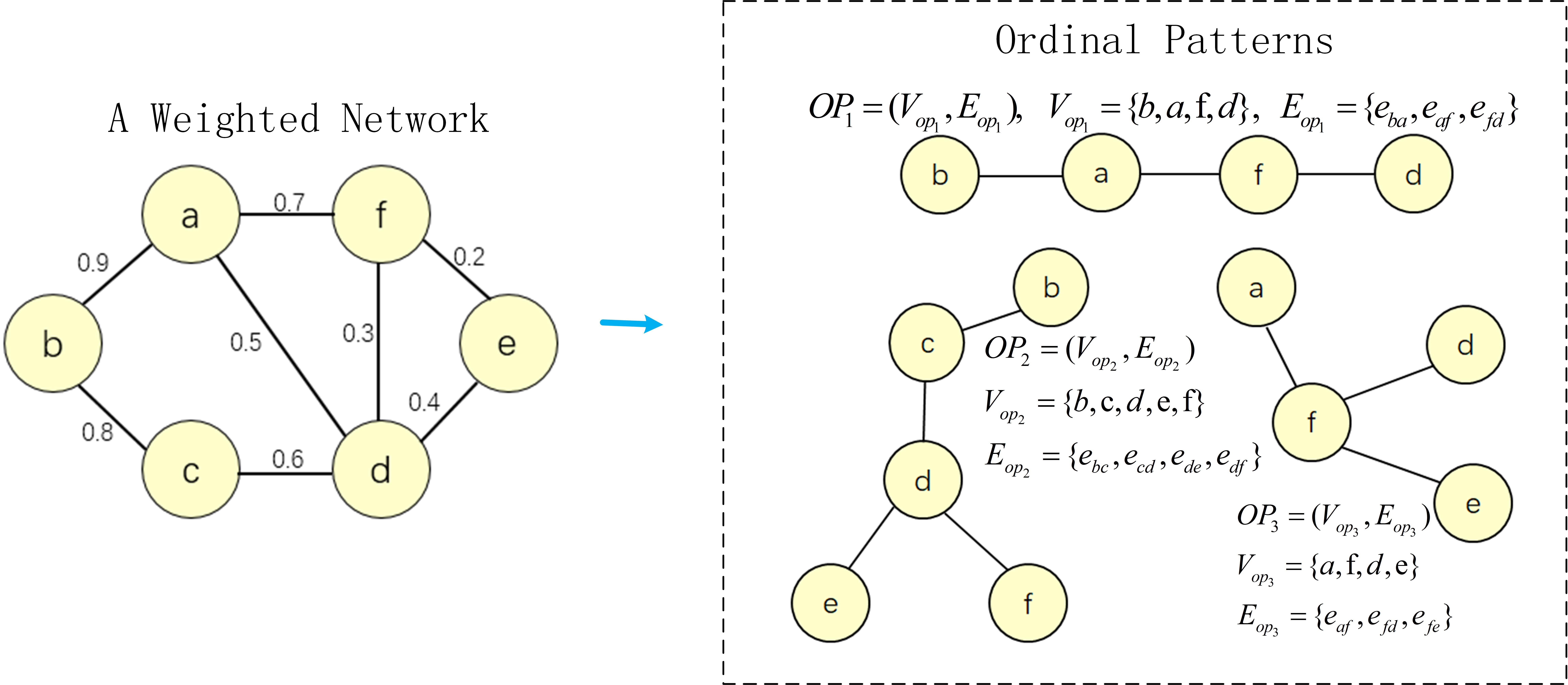}
\caption{Illustration of the ordinal patterns defined on a weighted network. Left is a weighted network, right is ordinal pattern.}
\label{Ordinal_pattern}
\end{figure}

A weighted network or graph \emph{G} consists of a set of nodes \emph{V}, edges \emph{E} and weight vectors \emph{W}, $\emph{G}=\left(\emph{V}, \emph{E}, \emph{W}\right)$. \emph{W} is the weight vector for those edges with the \emph{i}-th element $\emph{W}\left(e_i\right)$ representing the connection strength of the edge \emph{e}$_i$, $\emph{e}_i\in\emph{E}$. The ordinal pattern (\emph{OP}) defined in graph \emph{G} is a set including ordinal nodes and ordinal edges
$\emph{OP}=\left(\emph{V}_\emph{op}, \emph{E}_\emph{op}\right)$. $\emph{E}_\emph{op}$ is a ordinal edge set, $\emph{E}_\emph{op}=\{\emph{e}_1,\emph{e}_2,\cdots,\emph{e}_i,\emph{e}_j,\cdots,\emph{e}_M \}\subseteq\emph{E}$, all $0<\emph{i}<\emph{j}\leq\emph{M}$, $\emph{W}\left(\emph{e}_i\right)>\emph{W}\left(\emph{e}_j\right)$, $\emph{e}_i$ and $\emph{e}_j$ are called ordinal edges. $\emph{V}_\emph{op}$ is a vertex set where vertexes are in ordinal edges included in $\emph{E}_\emph{op}$.
The illustration of ordinal patterns could be seen in Figure~\ref{Ordinal_pattern}. \emph{OP$_1$}, \emph{OP$_2$} and \emph{OP$_3$} are ordinal patterns from a weighted network.

\subsection{Ordinal pattern isomorphism}
A graph or network could be decomposed into multiple ordinal patterns. The set consisting of all ordinal patterns is called ordinal pattern set $\left(\emph{OPs}\right)$. \emph{OPs}$_1$ and \emph{OPs}$_2$ are two ordinal pattern sets of graph \emph{G$_1$} and \emph{G$_2$},
\emph{OP$_1$}=$\left(\emph{V}_\emph{op$_1$},\emph{E}_\emph{op$_1$} \right)$ and \emph{OP$_2$}=$\left(\emph{V}_\emph{op$_2$},\emph{E}_\emph{op$_2$} \right)$
are two ordinal patterns, \emph{OP$_1$}$\in$\emph{OPs}$_1$, \emph{OP$_2$}$\in$\emph{OPs}$_2$.
An ordinal pattern isomorphism between two ordinal patterns \emph{OP$_1$} and \emph{OP$_2$} is a bijective mapping $\varphi: \emph{V}_\emph{op$_1$}\to\emph{V}_\emph{op$_2$}$, i.e. $\forall\emph{v}_\emph{op$_1$},\emph{v'}_\emph{op$_1$}\in\emph{V}_\emph{op$_1$}$: $(\emph{v}_\emph{op$_1$},\emph{v'}_\emph{op$_1$})\in\emph{E}_\emph{op$_1$}$ $\Leftrightarrow$ $\varphi\left(\emph{v}_\emph{op$_1$}\right),\varphi\left(\emph{v'}_\emph{op$_1$}\right)\in\emph{V}_\emph{op$_2$}$, $(\varphi\left(\emph{v}_\emph{op$_1$}\right),\varphi\left(\emph{v'}_\emph{op$_1$}\right))\in\emph{E}_\emph{op$_2$}$. \emph{OP$_1$} and \emph{OP$_2$} are isomorphic, written $\emph{OP$_1$}\cong\emph{OP$_2$}$.
%
%
\emph{V}$_1^{'}$ $\subseteq$ $\emph{V}_\emph{op$_1$}$ and \emph{V}$_2^{'}$ $\subseteq$ $\emph{V}_\emph{op$_2$}$ are subsets of ordinal pattern vertices. An ordinal pattern isomorphism $\tau$ of \emph{OP$_1$}[\emph{V}$_1^{'}$] and \emph{OP$_2$}[\emph{V}$_2^{'}$] is called sub-ordinal pattern isomorphism (\emph{SOPI}) of \emph{OP$_1$} and \emph{OP$_2$}.

\section{Ordinal pattern kernel}
\subsection{Ordinal pattern kernel}
We suppose that a graph \emph{G} has \emph{N} ordinal patterns, hence ordinal pattern set of graph \emph{G} is $\emph{OPs}=\{\emph{OP$_1$},\cdots,\emph{OP$_i$},\cdots,\emph{OP$_N$}\}$, \emph{OP$_i$} is the \emph{i-th} ordinal pattern, $1\leq\emph{i}\leq\emph{N}$. Two graphs \emph{G}$_1$ and \emph{G}$_2$ have their own \emph{OPs} respectively, they are \emph{OPs}$_1$ and \emph{OPs}$_2$. $\varphi_{ij}$ is the isomorphism mapping from ordinal pattern \emph{OP$_i$} to ordinal pattern \emph{OP$_j$},\emph{OP$_i$}$\in$\emph{OPs}$_1$, \emph{OP$_j$}$\in$\emph{OPs}$_2$. Let $\Psi\left(\emph{OP$_i$},\emph{OP$_j$}\right)$ refer to the set which includes all sub-ordinal pattern isomorphisms (\emph{SOPIs}) of \emph{OP$_i$}and \emph{OP$_j$} and $\Upsilon$: $\Psi\left(\emph{OP$_i$},\emph{OP$_j$}\right)\rightarrow \mathbb{R}^+$ a weight function. The sub-ordinal pattern isomorphism kernel is defined as:

\begin{equation}
\begin{aligned}
\emph{k}_{sopi}\left(\emph{OP$_i$},\emph{OP$_j$}\right)=\sum_{\tau\in\Psi\left(\emph{OP$_i$},\emph{OP$_j$}\right)}\Upsilon\left(\tau\right)
\end{aligned}
\end{equation}

\textbf{Theorem 1.} \emph{$k_{sopi}$ is positive semidefinite (p.s.d) kernel }

\emph{Proof}. Kernel \emph{$k_{sopi}$} counts the number of isomorphisms between sub-ordinal patterns in ordinal pattern \emph{OP$_i$} and \emph{OP$_j$}. We have known that the kernel counting the number of isomorphisms between graph sub-structures is $p.s.d$ [19]. In \emph{$k_{sopi}$}, sub-ordinal patterns could be treated as the special sub-structures where the nodes and edges are from ordinal patterns. Hence, \emph{$k_{sopi}$} is $p.s.d$. $\Box$


The sub-ordinal pattern isomorphism kernel measures the similarity between two ordinal patterns by counting the number of sub-ordinal pattern isomorphisms, $\bigtriangleup :=\emph{k}_{sopi}\left(\emph{OP$_i$},\emph{OP$_j$}\right)$. Then, we get the ordinal pattern kernel between two graphs \emph{G$_1$} and \emph{G$_2$}, defined as:
\begin{equation}
\begin{aligned}
\emph{k}_{op}\left(\emph{G$_1$},\emph{G$_2$}\right)=\sum_{\emph{OP$_i$}\in\emph{OPs$_1$}}\sum_{\emph{OP$_j$}\in\emph{OPs$_2$}}\emph{iso-count(OP$_i$,OP$_j$)}
\end{aligned}
\end{equation}
where \emph{OPs$_1$} and \emph{OPs$_2$} are the ordinal pattern sets of \emph{G}$_1$ and \emph{G}$_2$, \emph{iso-count($\cdot$)} is a function calculating isomorphism:
\begin{equation}
\emph{iso-count(OP$_i$,OP$_j$)}=\left\{
\begin{aligned}
&\lambda_{o}(OP_i) , ~if ~\emph{OP$_i$} ~and~ \emph{OP$_j$} ~are ~isomorphic \\
&\bigtriangleup , ~otherwise.
\end{aligned}
\right.
\end{equation}
where a weight function $\lambda_{o}$: $OPs\to\mathbb{R}^{+}$, which count the node number when two ordinal patterns are isomorphic.

\subsection{Ordinal pattern attribute kernel}
If the ordinal patterns have the node and edge attributes, the above formula is not appropriate to defining a mapping to preserve their attributes. We need to generalize the formula (2) to the common ordinal patterns with attributes. The new kernel is called ordinal pattern attribute (OPA) kernel.
\begin{equation}
\begin{aligned}
\emph{k}_{opa}\left(\emph{G$_1$},\emph{G$_2$}\right)=\sum_{\emph{OP$_i$}\in\emph{OPs$_1$}}\sum_{\emph{OP$_j$}\in\emph{OPs$_2$}}\emph{iso-count(OP$_i$,OP$_j$)}\cdot\emph{K$_{{V_{op}},{E_{op}}}$}
\end{aligned}
\end{equation}

\begin{equation}
\begin{aligned}
\emph{K$_{{V_{op}},{E_{op}}}$}=\prod_{\emph{v$_{op}$}\in\emph{V}_{\emph{op$_i$}} } \emph{K$_V$}\left(v_{op},\varphi\left(v_{op}\right)  \right) \prod_{\emph{e$_{op}$}\in\emph{E}_{\emph{op$_i$}} } \emph{K$_E$}\left(e_{op},\varphi\left(e_{op}\right)  \right)
\end{aligned}
\end{equation}
where \emph{K$_V$} and \emph{K$_E$} are two positive semidefinite kernel functions defined on ordinal pattern node and edge attribute features.

\textbf{Theorem 2.} \emph{The ordinal pattern attribute kernel is p.s.d}

Before we prove \emph{theorem 2}, we need to know \emph{R relation}  in ordinal pattern (seeing supplement) and \emph{R-ordinal pattern convolution}. Suppose \emph{OPs$_1$} and \emph{OPs$_1$} are two ordinal pattern set $\left(OPs\right)$ in graph \emph{G$_1$} and \emph{G$_2$}. Ordinal pattern $OP\in\emph{OPs$_1$}$ and $OP'\in\emph{OPs$_2$}$, $OP=\left(V_{op},E_{op} \right)$, $OP'=\left(V_{op'},E_{op'} \right)$. The decompositions of $OP$ and $OP'$: $\vec{OP}=OP_1,\cdots,OP_M$ and $\vec{OP'}=OP'_1,\cdots,OP'_M$ are two parts of $OP$ and $OP'$. For $1\leq i\leq M$, we define a function \emph{iso-count$_i$($OP_i$,$OP'_i$)} on $OP_i$ and $OP'_i$ that could be used to measure the similarity of ordinal patterns $OP_i$ and $OP'_i$. For $OP_i=\left(V_{op_i},E_{op_i} \right)$, $OP'_i=\left(V_{op'_i},E_{op'_i} \right)$, we define two positive semidefinite kernels $K_{V_i}\left(v_{op_i},v_{op'_i} \right)$ and $K_{E_i}\left(e_{op_i},e_{op'_i} \right)$ that could be used to measure the similarity of the ordinal nodes and edges, $v_{op_i}\in V_{op_i}$, $v_{op'_i}\in V_{op'_i}$, $e_{op_i}\in E_{op_i}$, $e_{op'_i}\in E_{op'_i}$. Then we define the kernel $K\left(OPs_1,OPs_2 \right)$ measuring the similarity between ordinal pattern set \emph{OPs$_1$} and \emph{OPs$_2$} as the following ordinal pattern convolution.
\begin{equation}
\begin{aligned}
K\left(OPs_1,OPs_2 \right)=\sum_{\vec{OP}\in R^{-1}\left(OP \right)}\sum_{\vec{OP'}\in R^{-1}\left(OP' \right)}\emph{iso-count$_i$($OP_i$,$OP'_i$)}\cdot\emph{K$_{{V_i},{E_i}}$}
\end{aligned}
\end{equation}

\begin{equation}
\begin{aligned}
K_{{V_i},{E_i}}=\prod^M_{i=1}K_{V_i}\left(v_{op_i},v_{op'_i} \right) \prod^M_{i=1} K_{E_i}\left(e_{op_i},e_{op'_i} \right)
\end{aligned}
\end{equation}
where $K\left(OPs_1,OPs_2 \right)$ is a symmetric function on $\mathbb{Q}\times\mathbb{Q}$, $\mathbb{Q}=\{OP:R^{-1}\left(OP\right) is ~nonempty\}$. \emph{K} is kernel defined on \emph{iso-count$_i$($OP_i$,$OP'_i$)}, $K_{V_i}\left(v_{op_i},v_{op'_i} \right)$ and $K_{E_i}\left(e_{op_i},e_{op'_i} \right)$ by \emph{R} relation. Hence \emph{K} is called \emph{R-ordinal pattern convolution} which is the zero extension of \emph{K} to $OPs\times OPs$. If \emph{R} is finite, then \emph{K} is a finite convolution.

\textbf{Theorem 3.} \emph{R-ordinal pattern convolution is a kernel on $OPs\times OPs$}

\emph{Proof}. Let \emph{U} denote $SOPs_1\times\cdots\times SOPs_M$, $SOPs_i$ is ordinal pattern subset, $1\leq i\leq M$, since \emph{iso-count$_i$($OP_i$,$OP'_i$)}, $K_{V_i}\left(v_{op_i},v_{op'_i} \right)$ and $K_{E_i}\left(e_{op_i},e_{op'_i} \right)$ are kernels. We know that $\bar{K}\left(\vec{OP},\vec{OP'} \right)$ is a kernel on $U\times U$, obviously it is the kernel closure under tensor product
\begin{equation}
\begin{aligned}
\bar{K}\left(\vec{OP},\vec{OP'} \right)=\emph{iso-count$_i$($OP_i$,$OP'_i$)}\cdot\prod^M_{i=1}K_{V_i}\left(v_{op_i},v_{op'_i} \right) \prod^M_{i=1} K_{E_i}\left(e_{op_i},e_{op'_i} \right)
\end{aligned}
\end{equation}
Since \emph{R} is finite, according to Lemma 1, $\bar{K'}\left(R^{-1}\left(OP\right),R^{-1}\left(OP'\right) \right)$ is a kernel on the product of the set of all not empty $R^{-1}\left(OP\right)$ such that $OP\in OPs$ with itself.
\begin{equation}
\begin{split}
\bar{K'}\left(R^{-1}\left(OP\right),R^{-1}\left(OP'\right) \right)&=\sum_{\vec{OP}\in R^{-1}\left(OP \right)}\sum_{\vec{OP'}\in R^{-1}\left(OP' \right)}\bar{K}\left(\vec{OP},\vec{OP'} \right)\\
&=\sum_{\vec{OP}\in R^{-1}\left(OP \right)}\sum_{\vec{OP'}\in R^{-1}\left(OP' \right)}\emph{iso-count$_i$($OP_i$,$OP'_i$)}\cdot\emph{K$_{{V_i},{E_i}}$}
\end{split}
\end{equation}
Since \emph{R-ordinal pattern convolution} is the zero extension of $K\left(OPs_1,OPs_2 \right)=\bar{K'}\left(R^{-1}\left(OP\right),R^{-1}\left(OP'\right) \right)$, it follows that it is a kernel on $OPs\times OPs$. The OPA kernel is a \emph{R-ordinal pattern convolution kernel} , where each kernel is \emph{p.s.d}, hence \emph{OPA kernel is p.s.d}.$\Box$

\textbf{Lemma 1} If \emph{K} is a kernel on a set $U\times U$, for all nonempty and finite set $X,Y\subseteq U$. We define a function $K'\left(X,Y \right)=\sum_{x\in X, y\in Y}K\left(x,y \right)$. Then $K'$ is a kernel on the product of the set of all nonempty and finite subsets of \emph{U} with itself.

\emph{Proof}. For any finite nonempty subset $X\subseteq U$, let $f_X=\sum_{u\in X}K_u\in H_0$, where $H_0$ is the pre-Hilbert space associated with $K'$. If for nonempty finite $X,Y\subseteq U$, we define $K'\left(X,Y \right)=\sum_{u\in X, v\in Y}K\left(u,v \right)$, then by Equation (1) in supplement, $K'\left(X,Y \right)=\langle f_X,f_Y\rangle $. Because an inner product is a kernel, it follows that $K'$ is a kernel on the product of the set of all nonempty finite subsets of \emph{U} with itself.$\Box$

Ordinal pattern kernel in formula (2) is a special case of ordinal pattern attribute kernel in formula (4). When we do not take node and edge attributes into the computation of ordinal pattern kernel, the kernel on node and edge attributes is equal 1, $\emph{K$_{{V_{op}},{E_{op}}}$}=1$, hence formula (4) degenerate to formula (2).



These kernels guarantee that exactly the conditions of ordinal pattern isomorphism are fulfilled. Hence, the $OP$ kernel is a special case of OPA kernel, we could obtain the following corollary.

\textbf{Corollary 1.} \emph{The ordinal pattern kernel is p.s.d}

Although we have known how to calculate ordinal pattern kernel and ordinal pattern attribute kernel, there is another problem in them. The problem is that computing the kernels including ordinal pattern kernel and ordinal pattern attribute kernel based on ordinal pattern isomorphism are NP-hard.

\textbf{Theorem 4.} \emph{Computing the kernel based on ordinal pattern isomorphism is NP-hard.}

\emph{Proof}. Let \emph{OP$_n$}$\in$\emph{OPs} be the ordinal pattern with \emph{n} edges and let \textbf{e}$_{op}$ be a vector in the ordinal pattern feature space which is defined by the mapping $\phi$: \emph{OPs} $\rightarrow$ $\mathcal{H}$ into Hilbert space $\mathcal{H}$ with one feature $\phi_{OP}$ for each ordinal pattern $OP\in OPs$, $OP'\in OPs$, $\phi_{OP}=\lambda_{|\varepsilon\left(\emph{OP} \right)|}|\{OP'\in OPs: OP'\cong OP \}|$,
where $\lambda$ is a sequence $\lambda_1$, $\lambda_2$,$\cdots$, $\lambda_N$ of weights $\left(\lambda_i\in\mathbb{R}; \lambda_i>0 ~for~ all~ i\in\emph{N} \right)$. In \textbf{e}$_{op}$, the features corresponding to \emph{OP} equal 1 and others equal 0. Let $OP'$ be any ordinal pattern with \emph{m} vertices. According to [11] for linear independent \{$\phi$$\left(\emph{OP$_n$} \right)$ \}$_{n\in\emph{N}}$, we could always find $\alpha_1$,$\cdots$,$\alpha_m$ in polynomial time, make that $\alpha_1\phi\left(OP_1 \right)$+,$\cdots$,+$\alpha_m\phi\left(OP_m\right)$=\textbf{e}$_{OP_m}$. Then
$\alpha_1\langle\phi\left(OP_1 \right)$,$\phi\left(OP'\right)\rangle$+,$\cdots$,+$\alpha_m\langle\phi\left(OP_m \right)$,$\phi\left(OP'\right)\rangle$ >0 if and only if $OP'$ has a Hamiltonian path. We all known that finding a Hamiltonian path is \emph{NP}-complete. Hence, computing the kernel based on ordinal pattern isomorphism is NP-hard. $\Box$

\section{Modified ordinal pattern kernel}
There are two problems in above computation. One problem is that the ordinal pattern has Hamiltonian path which is \emph{NP}-complete problem. The other problem is that ordinal pattern \emph{OP$_1$} is the sub-ordinal pattern of another ordinal pattern \emph{OP$_2$}. For example, in Figure~\ref{Ordinal_pattern}, $\emph{OP$_1$}=\left(\emph{V$_{op_1}$},\emph{E$_{op_1}$} \right)$, \emph{V$_{op_1}$}=\{a,b,c\}, \emph{E$_{op_1}$}=\{\emph{e$_{ab}$}, \emph{e$_{bc}$}\},
$\emph{OP$_2$}=\left(\emph{V$_{op_2}$},\emph{E$_{op_2}$} \right)$, \emph{V$_{op_2}$}=\{a,b,c,d\}, \emph{E$_{op_2}$}=\{\emph{e$_{ab}$}, \emph{e$_{bc}$}, \emph{e$_{cd}$}\}, $\emph{OP$_1$} ~is~ the~ sub-structure~ of~ $\emph{OP$_2$}. This problem brings redundant calculations for ordinal pattern kernel. In order to overcome these two problems, we propose a modified ordinal pattern kernel based on depth-first search. We adopt the depth-first search (DFS) algorithm to seek the deepest ordinal pattern for each node in the graph and then design the relevant ordinal pattern kernel on them. The ordinal pattern constructed by depth-first search is called depth-first-based ordinal pattern (DOP). The DOP is a linear structure, hence the isomorphism problem in DOP could be regarded as a matched problem.

\subsection{Depth-first-based ordinal pattern (\emph{DOP})}
Here, we detect the establishment process of depth-first-based ordinal pattern in detail. A weighted network or graph \emph{G} consists of a set of nodes \emph{V}, edges \emph{E} and weight vectors \emph{W}, $\emph{G}=\left(\emph{V},\emph{E},\emph{W}\right)$. $\forall u\in\emph{V}$, the neighborhood vertex set of a vertex \emph{u}: $\delta\left(\emph{u}\right)=\{v:\left(u,v\right)\in E, v\in V\}$,  the edge weight set between vertex \emph{u} and its neighborhood vertexes: $\mathbb{W}\left(u\right)=\{\emph{W}\left(u,v\right): v\in \delta\left(\emph{u}\right), \left(u,v\right)\in E \}$. In graph \emph{G}, we arbitrarily choose a node as the start node \emph{v$_0$} and use the depth-first search algorithm to seek the deepest ordinal pattern of node \emph{v$_0$}. The detailed process of constructing depth-first-based ordinal pattern for node \emph{v$_0$} could be seen in \textbf{Algorithm 1}.

\begin{algorithm}
\caption{\emph{DOP(W,v$_0$)}} 
\hspace*{0.02in} {\bf Input:} 
Weight matrix \emph{W} of graph \emph{G}, start node v$_0$\\
\hspace*{0.02in} {\bf Output:} 
The deepest ordinal pattern of node v$_0$
\begin{algorithmic}[1]

\State Visit node v$_0$ mark it as a visited node.\\

Select a non-visited node v$_1$ from $\delta\left(\emph{v$_0$}\right)$ to visit and mark it as a a visited node, node v$_1$ needs to meet the condition: $\emph{W}\left(v_0,v_1 \right)=max\{\mathbb{W}\left(v_0\right)\}$\\

Choose a non-visited node v$_2$ from $\delta\left(\emph{v$_1$}\right)$ to visit and mark it, node v$_2$ needs to meet the condition: $\emph{W}\left(v_1,v_2 \right)=max\{\mathbb{W}\left(v_1\right)\}$ and $\emph{W}\left(v_1,v_2 \right)<\emph{W}\left(v_0,v_1 \right)$. If $\emph{W}\left(v_1,v_2 \right)\geq\emph{W}\left(v_0,v_1 \right)$, select another node $v'_2$ from $\delta\left(\emph{v$_1$}\right)$ to replace v$_2$, make node $v'_2$ meet the condition: $\emph{W}\left(v_1,v'_2\right)=max\{\mathbb{W}\left(v_1\right)-\emph{W}\left(v_1,v_2\right)\}$ and $\emph{W}\left(v_1,v'_2 \right)<\emph{W}\left(v_0,v_1 \right)$. The node v$_1$ and v$_2$ are called ordinal nodes, $\left(v_0,v_1 \right)$ and $\left(v_1,v_2\right)$ are called ordinal edges.\\

Repeat step 3 and seek the next ordinal node of each visited node in turn, when the visited node does not have the next ordinal node, the search process is stopped.\\

Output the deepest ordinal pattern of start node v$_0$.

\end{algorithmic}

\end{algorithm}

\subsection{Depth-first-based ordinal pattern attribute kernel }
We could use Algorithm 1 to calculate the DOP for each node in graph \emph{G}. Subsequently, we could utilize these ordinal patterns to construct the depth-first-based ordinal pattern kernel \emph{k$_{DOP}$} between graph \emph{G$_1$} and \emph{G$_2$} with attributes. Because ordinal pattern kernel is a special case of ordinal pattern attribute kernel. Here, we only detect the depth-first-based ordinal pattern attribute kernel.
\begin{equation}
\begin{aligned}
\emph{k$_{DOP}$}\left(\emph{G$_1$},\emph{G$_2$}\right)=\sum_{\emph{v$_i$}\in\emph{G$_1$}}\sum_{\emph{v$_j$}\in\emph{G$_2$}}match\left(DOP\left(v_i\right),
DOP\left(v_j\right) \right)\cdot\emph{K$_{{V_{DOP}},{E_{DOP}}}$}
\end{aligned}
\end{equation}

\begin{equation}
\begin{aligned}
\emph{K$_{{V_{DOP}},{E_{DOP}}}$}=\prod_{\substack{\emph{v$_{DOP}$}\in\emph{V}_{DOP\left(v_i\right)}\\ \emph{v'$_{DOP}$}\in\emph{V}_{DOP\left(v_j\right)}  } } \emph{K$_V$}\left(v_{DOP},v'_{DOP}  \right) \prod_{\substack{\emph{e$_{DOP}$}\in\emph{E}_{DOP\left(v_i\right)}\\ \emph{e'$_{DOP}$}\in\emph{E}_{DOP\left(v_j\right)}  } } \emph{K$_E$}\left(e_{DOP},e'_{DOP}  \right)
\end{aligned}
\end{equation}
where \emph{DOP$\left(v_i\right)$} and \emph{DOP$\left(v_j\right)$} are the depth-first-based ordinal patterns or the deepest ordinal patterns of node $v_i$ and $v_j$. $\emph{K$_V$}\left(v_{DOP},v'_{DOP}  \right)$ and $\emph{K$_E$}\left(e_{DOP},e'_{DOP}  \right)$ are the kernels defined on node and edge attributes in \emph{DOP$\left(v_i\right)$} and \emph{DOP$\left(v_j\right)$}.

\begin{equation}
\begin{aligned}
match\left(DOP\left(v_i\right),DOP\left(v_j\right) \right)=\sum_{p\subseteq DOP\left(v_j\right) }\lambda_{|V\left(p \right)|}|\{ q\subseteq DOP\left(v_i\right): p\cong q\}|
\end{aligned}
\end{equation}
where $\lambda$ is a sequence $\lambda_1$,$\lambda_2$,$\cdots$,$\lambda_N$ of weights $\left(\lambda_i\in\mathbb{R}; \lambda_i>0 ~for~ all~ i\in N\right)$. Obviously, formula $\left(10\right)$ is a special case of formula $\left(4\right)$. Here, $match\left(DOP\left(v_i\right),DOP\left(v_j\right) \right)$ could also be calculated by the matched node numbers between $DOP\left(v_i\right)$ and $DOP\left(v_j\right)$.

\textbf{Theorem 5.} \emph{Depth-first-based ordinal patternn attribute kernel is p.s.d.}

\emph{Proof}. According to the definitions of the sub-ordinal pattern isomorphisms and the corresponding sub-ordinal pattern isomorphism kernel, we know that $match\left(DOP\left(v_i\right),DOP\left(v_j\right) \right)$ is \emph{p.s.d}. \emph{K$_{{V_{DOP}},{E_{DOP}}}$} is a \emph{p.s.d} kernel defined in node and edge attributes. $\emph{k$_{DOP}$}\left(\emph{G$_1$},\emph{G$_2$}\right)$ is a \emph{R-ordinal pattern convolution kernel}, hence depth-first-based ordinal pattern attribute kernel is \emph{p.s.d}. $\Box$

\textbf{Kernel selection.}  In real applications, such as brain neuroimaging, each brain structure is abstracted into a network (or graph) where each node and edge may have multi-dimensional attributes. These attributes are from Euclidean spaces, we could utilize the Gaussian RBF kernel or linear kernel to compute the kernel \emph{K$_V$} and \emph{K$_E$} in node and edge attributes. If the attributes are discrete, we use Delta kernel $k_{attribute}\left(a,b\right)=\mathcal{T}_{\{a=b\}}$ [12].

\section{Experiments}
In this section, we perform the classification, robustness and discriminative sub-structure mining experiments in the brain  network data of brain disease patients and normal controls to verify the effectiveness of ordinal pattern kernel. All the experiments are performed on a server with an Intel$\left(R\right)$ Core$\left(TM\right)$ $i7-8700$, $3.20GHz$ CPU and 32GB RAM having 6 cores and 12 threads.

\begin{table}
\centering
\caption{Classification results among EMCI, LMCI, AD and NC (\%)}
\label{classification}
\resizebox{\textwidth}{20mm}{
\begin{tabular}{ccccccc}\hline \toprule
Graph kernel method & MCI vs. NC  & AD vs. NC & EMCI vs. LMCI & EMCI vs. AD & LMCI vs. AD \\\hline
SP      &67.79 &76.19 &56.57 &76.67 &77.92 \\
WL-ST   &67.79 &75.00 &73.74 &73.33 &71.43 \\
WL-SP   &75.83 &77.38 &74.75 &77.78 &72.73   \\
RW      &66.44 &73.80 &72.73 &76.67 &77.92   \\
PM      &79.87 &72.62 &75.76 &74.44 &75.32   \\
WWL     &73.83 &67.86 &76.77 &71.11 &80.52   \\
GH      &71.81 &64.29 &56.57 &65.56 &61.04   \\
Tree++  &74.50 &60.71 &57.58 &65.56 &63.64   \\
\textbf{DOP}     &\textbf{81.21}	&\textbf{86.90}	&\textbf{83.84}	&\textbf{80.0}	&\textbf{83.12}  \\\hline \toprule
\end{tabular}}
\end{table}
\subsection{Datasets}
The brain network data used in the experiments are based on brain disease patients and normal controls, which are constructed from the resting state functional magnetic resonance imaging (fMRI) data deriving from the online database ADNI\footnote{http://adni.loni.usc.edu/}. Brain diseases are Alzheimer’s disease (AD), early mild cognitive impairment (EMCI), late mild cognitive impairment (LMCI). MCI consists of EMCI and LMCI. The  normal controls (NC) matched to brain disease patients are used to classify brain diseases and seek the discriminative sub-structures from brain disease networks. The fMRI data are preprocessed with statistical parametric mapping (SPM) \footnote{http://www.fil.ion.ucl.ac.uk/spm} and resting-state fMRI analysis toolkit (REST) \footnote{http://www.restfmri.net}. The detailed fMRI data preprocessing steps could be seen in supplement.

\subsection{Brain network construction}
After processing the fMRI data of brain disease patients and normal controls, we need to transform the preprocessed fMRI data into brain networks. The whole-brain cortical and subcortical structures are subdivided into 90 brain regions for each subject based on the AAL atlas. The linear correlation between mean time series of a pair of brain regions is then calculated to measure the functional connectivity. At last, a $90\times90$ fully-connected weighted functional network is constructed for each subject. The detailed contents of constructing brain network could be seen in supplement.

\subsection{Experimental setup}
In the experiments, uniform weight $\lambda$ is chosen from $\{10^{-2},10^{-1},\cdots,10^2\}$. We compare our kernel with state-of-the-art graph kernels including shortest path kernel (SP) [9], Weisfeiler-Lehman subtree kernel (WL-ST) [17], Weisfeiler-Lehman shortest path kernel (WL-SP) [17], random walk kernel (RW) [10], pyramid match kernel (PM) [21], Wasserstein Weisfeiler-Lehman graph kernel (WWL) [18], GraphHopper kernel (GH) [23], Truncated Tree Based Graph Kernels (Tree++) [24]. In robustness experiment, we randomly discard partial data in each brain network by 25\% missing rate.

Support Vector Machine (SVM) [25] as our final classifier is exploited to conduct the classification experiment. We perform leave-one-out cross-validation for all the classifications, using one sample for testing and the others for training. The tradeoff parameter \emph{C} in the SVM is selected from $\{10^{-3},10^{-2},\cdots,10^3\}$.

\subsection{Results and discussion}
The classification accuracy, method robustness and discriminative sub-structures are very important in brain network analysis. We report the classification accuracies in Table~\ref{classification} and investigate the robustness of our depth-first-based ordinal pattern kernel in the special missing rate in Figure~\ref{Missingdata} and the discriminative ordinal patterns in Figure~\ref{Discriminative}.

\subsubsection{Classification results}
We conduct the classification experiments in the brain functional networks of brain disease patients and normal controls, including EMCI, LMCI, AD and NC matched to brain diseases. The MCI consists of EMCI and LMCI. We compare our ordinal pattern kernel with the state-of-the-art graph kernels in these datasets, seeing Table~\ref{classification}. In these datasets, our method outperforms the state-of-the-art graph kernels.

\subsubsection{Robustness}
\begin{figure}[h]
\setlength{\abovecaptionskip}{0.cm}
\centering
\includegraphics[width=14cm,height=3.4cm]{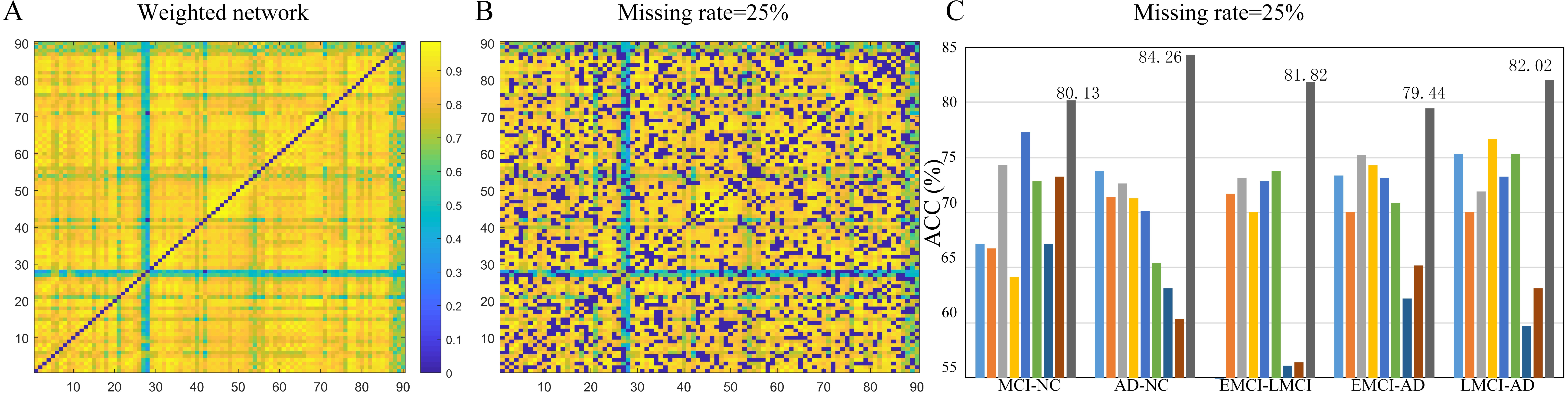}
\caption{Classification accuracy in network data having 25\% missing rate. }
\label{Missingdata}
\end{figure}

The missing information usually exists in the brain fMRI data [26, 27], which will bring challenges for brain network classification. Here, we randomly discard the specific percentage of data in each brain network, seeing Figure~\ref{Missingdata} (A)-(B). The specific percentage is 25\%. Then, we measure the similarities among these specific brain networks with our ordinal pattern kernel. The experimental results indicate that our ordinal pattern kernel could acquire robust and excellent classification accuracies in these missing brain network data, seeing Figure~\ref{Missingdata} (C).
\subsubsection{Discriminative sub-structures}
\begin{figure}[h]
\setlength{\abovecaptionskip}{0.cm}
\centering
\includegraphics[width=14cm,height=3cm]{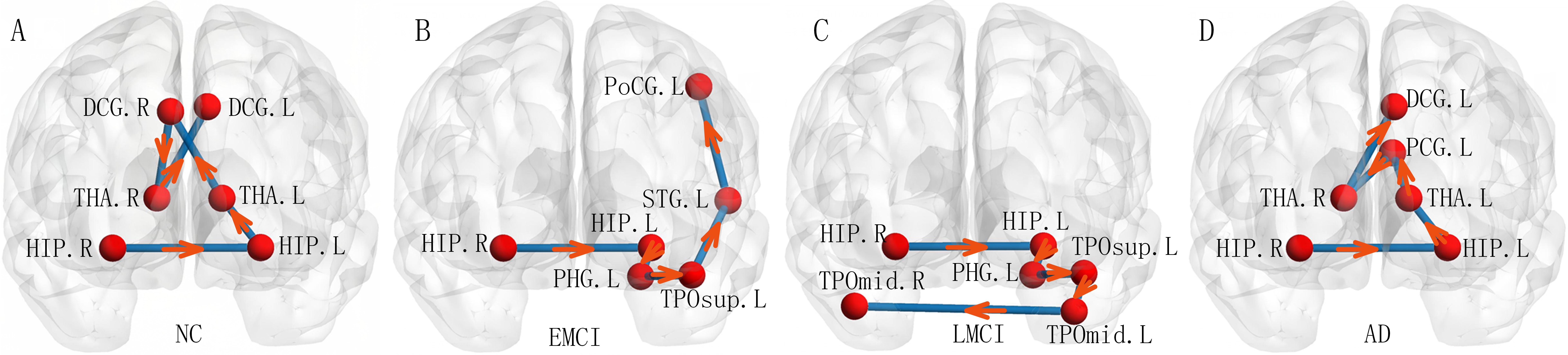}
\caption{ Discriminative ordinal patterns from right hippocampus}
\label{Discriminative}
\end{figure}

We could also use our proposed depth-first-based ordinal pattern to seek the discriminative sub-structures [28] in brain networks of patients and normal controls. We plot the top six discriminative ordinal patterns identified in the classification tasks in Figure~\ref{Discriminative} (A)-(D). The start node of depth-first-based ordinal pattern is right hippocampus.

We could find that the depth-first-based ordinal patterns in brain diseases are different from those in normal controls. The first two ordinal nodes in depth-first-based ordinal patterns are same among EMCI, LMCI, AD and NC. The first four ordinal nodes in EMCI are same to those in LMCI, seeing Figure~\ref{Discriminative} (B) and (C). From Figure~\ref{Discriminative} (B)-(D), we could also find that ordinal pattern structures are gradually changed from EMCI to AD. The top discriminative depth-first-based ordinal patterns starting from left hippocampus could be seen in supplement.


\section{Conclusion}
In this paper, we propose the ordinal pattern kernel for making full use of ordinal edge weight information to measure the similarities between brain networks.  We perform the classification, robustness and discriminative sub-structure mining experiments in the brain network data of brain disease patients and normal controls. The classification results indicate that our method outperforms the existing state-of-the-art graph kernels in the accuracy and robustness. Our proposed ordinal patterns based on depth-first search could capture the discriminative ordinal sub-structures in the brain networks of EMCI, LMCI and AD.






\section*{Broader Impact}
Ordinal pattern kernel is a method measuring the similarities between brain networks, which could be used to diagnose the brain diseases. This method could also be extended to the other networks having weight information. This work uses the public datasets from Alzheimer's Disease Neuroimaging Initiative (ADNI).

\section*{References}
\small
[1] ShuaiZong Si, XiaoLiu, JinFa Wang, et al. Brain networks modeling for studying the mechanism underlying the development of Alzheimer’s disease. Neural Regeneration Research, 14(10):1805, 2019.

[2] Amir Hossein Ghaderi, Mohammad Ali Nazari, Hassan Shahrokhi, et al. Functional Brain Connectivity Differences Between Different ADHD Presentations: Impaired Functional Segregation in ADHD-Combined Presentation but not in ADHD-Inattentive Presentation. Basic \& Clinical Neuroscience,8(4):267-278, 2017.

[3] Corey Fee, Mounira Banasr, Etienne Sibille. Somatostatin-Positive Gamma-Aminobutyric Acid Interneuron Deficits in Depression: Cortical Microcircuit and Therapeutic Perspectives. Biological psychiatry, 82(8):549-559, 2017.

[4] Qingbao Yu, Jing Sui,  Kent A. Kiehl, et al. State-related functional integration and functional segregation brain networks in schizophrenia. Schizophrenia Research, 150(2–3):450–458, 2013.

[5] Mikail Rubinov, Olaf Sporns.  Complex network measures of brain connectivity: Uses and interpretations. Neuroimage 52(3):1059–1069, 2010.

[6] Kai Ma, Jintai Yu, Wei Shao, et al. Functional Overlaps Exist in Neurological and Psychiatric Disorders: A Proof from Brain Network Analysis. Neuroscience 425:39–48, 2020.

[7] Christopher Morris, Nils M. Kriege, Kristian Kersting, et al. Faster kernels for graphs with continuous attributes via hashing. In Proceedings of the 16th IEEE International Conference on Data Mining (ICDM), 1095–1100, 2016.

[8] Yu Tian, Long Zhao, Xi Peng, et al. Rethinking Kernel Methods for Node Representation Learning on Graphs. In Advances in Neural Information Processing Systems (NeurIPS), 11686-11697, 2019.

[9] Karsten M. Borgwardt, Hanspeter Kriegel. Shortest-path kernels on graphs. Fifth IEEE International Conference on Data Mining (ICDM), 74-81, 2005.

[10] S V N Vishwanathan, Nicol N. Schraudolph, Risi Kondor, et al. Graph kernels. Journal of Machine Learning Research (JMLR), 11(Apr):1201–1242, 2010.

[11] Thomas Gartner, Peter Flach, Stefan Wrobel. On graph kernels: Hardness results and efficient alternatives. In Learning Theory and Kernel Machines, 129–143, 2003.

[12] Zhen Zhang, Mianzhi Wang, Yijian Xiang, et al. RetGK: Graph kernels based on Return Probabilities of Random Walks. In Advances in Neural Information Processing Systems (NeurIPS), 3964-3974, 2018.

[13] Tamas Horvath, Thomas Gartner, Stefan Wrobel. Cyclic Pattern Kernels for Predictive Graph Mining.  Advances in Knowledge Discovery \& Data Mining (KDD), 158-167, 2004.

[14] Tamas Horvath. Cyclic Pattern Kernels Revisited. Advances in Knowledge Discovery \& Data Mining (KDD), 791-801, 2005.

[15] Pierre Mahe, Jeanphilippe Vert. Graph kernels based on tree patterns for molecules. Machine Learning, 75(1):3-35, 2009.

[16] Giovanni Da San Martino, Nicolò Navarin, Alessandro Sperduti. Tree-based kernel for graphs with continuous attributes. IEEE transactions on neural networks and learning systems (TNNLS), 29(7):3270-3276, 2018.

[17] Nino Shervashidze, Pascal Schweitzer, Erik Jan van Leeuwen, et al. Weisfeiler-lehman graph kernels. Journal of Machine Learning Research (JMLR),12(Sep):2539–2561, 2011.

[18] Matteo Togninalli, Elisabetta Ghisu, Felipe Llinares Lopez, et al. Wasserstein Weisfeiler-lehman graph kernels. In Advances in Neural Information Processing Systems (NeurIPS), 6439-6449, 2019.

[19] Nils M. Kriege, Petra Mutzel. Subgraph matching kernels for attributed graphs. International Conference on Machine Learning (ICML), 291-298, 2012.

[20] Kristen Grauman, Trevor Darrell. The Pyramid Match Kernel: Efficient Learning with Sets of Features. Journal of Machine Learning Research (JMLR), 725-760, 2007.

[21] Giannis Nikolentzos, Polykarpos Meladianos, Michalis Vazirgiannis. Matching Node Embeddings for Graph Similarity. Association for the Advance of Artificial Intelligence (AAAI), 2429-2435, 2017.

[22] Daoqiang Zhang, Jiashuang Huang, Biao Jie, et al. Ordinal Pattern: A New Descriptor for Brain Connectivity Networks. IEEE Transactions on Medical Imaging, 37(7):1711-1722, 2018.

[23] Aasa Feragen, Niklas Kasenburg, Jens Petersen, et al. Scalable kernels for graphs with continuous attributes. In Advances in Neural Information Processing Systems (NeurIPS), 216-224, 2013.

[24] Wei Ye, Zhen Wang, Rachel Redberg, et al. Tree++: Truncated Tree Based Graph Kernels. IEEE Transactions on Knowledge and Data Engineering (TKDE), 2019.

[25] Chihchung Chang, Chihjen Lin. LIBSVM: A library for support vector machines. Acm Transactions on Intelligent Systems \& Technology, 2011.

[26] Hien M Nguyen, Gary H Glover. A Modified Generalized Series Approach: Application to Sparsely Sampled fMRI. IEEE Transactions on Biomedical Engineering, 60(10): 2867-2877, 2013.

[27] Jutta de Jong, Serge Dumoulin, Barrie Klein, et al. Hand position modulates visually-driven fMRI responses in premotor cortex. Society for Neuroscience. 2016.

[28] Fei Fei, Biao Jie, Daoqiang Zhang. Frequent and Discriminative Subnetwork Mining for Mild Cognitive Impairment Classification. Brain Connectivity, 4(5):347-360, 2014.

\end{document}